\RequirePackage{amsmath}
\documentclass[runningheads]{llncs}
\usepackage{graphicx} 
\usepackage{amsmath}
\usepackage{ifthen}
\usepackage{adjustbox}
\usepackage{xkeyval}
\usepackage{tikz}
\usepackage{calc}
\usepackage{amssymb}
\usepackage{multirow}
\usepackage{paralist}
\usepackage{comment}
\usepackage[utf8]{inputenc} 
\usepackage[T1]{fontenc}    
\usepackage{url}            
\usepackage{booktabs}       
\usepackage{xcolor}         
\usepackage{arydshln}
\usepackage{wrapfig}
\usepackage{bm}
\usepackage{xspace}
\usepackage{nicefrac}       
\usepackage{microtype}      
\usepackage[hidelinks]{hyperref}
\usepackage{amsmath,amssymb} 
\usepackage[normalem]{ulem}
\useunder{\uline}{\ul}{}
\usepackage{colortbl}
\usepackage[normalem]{ulem}

\def\etal{\emph{et al}. }
\def\eg{\emph{e}.\emph{g}.}
\def\ie{\emph{i}.\emph{e}.}

\usepackage[capitalize]{cleveref}
\crefname{section}{Sec.}{Secs.}
\Crefname{section}{Section}{Sections}
\Crefname{table}{Table}{Tables}
\crefname{table}{Tab.}{Tabs.}

\newcommand{\mysection}[1]{\vspace{-0.0mm}\section{#1}\vspace{-0.0mm}}
\newcommand{\mysubsection}[1]{\vspace{-0.0mm}\subsection{#1}\vspace{-0.0mm}}

\newcommand{\diff}[1]{\footnotesize{\textcolor{gray}{~($#1$)}}}

\begin{document}

\title{CLEFT: Language-Image Contrastive Learning with Efficient Large Language Model and Prompt Fine-Tuning}

\author{
Yuexi~Du\inst{1},~
Brian~Chang\inst{1},~
Nicha~C.~Dvornek\inst{1,2} 
}
\institute{Department of Biomedical Engineering, 
\and Department of Radiology \& Biomedical Imaging, \\Yale University, New Haven, CT, USA \\
\email{\{yuexi.du, brian.chang, nicha.dvornek\}@yale.edu}
}

\authorrunning{Y. Du et al.}

\titlerunning{CLEFT}

\maketitle

\begin{abstract}
    Recent advancements in Contrastive Language-Image Pre-training (CLIP)~\cite{radford2021learning} have demonstrated notable success in self-supervised representation learning across various tasks. However, the existing CLIP-like approaches often demand extensive GPU resources and prolonged training times due to the considerable size of the model and dataset, making them poor for medical applications, in which large datasets are not always common. Meanwhile, the language model prompts are mainly manually derived from labels tied to images, potentially overlooking the richness of information within training samples. 
    We introduce a novel language-image Contrastive Learning method with an Efficient large language model and prompt Fine-Tuning (CLEFT) that harnesses the strengths of the extensive pre-trained language and visual models. Furthermore, we present an efficient strategy for learning context-based prompts that mitigates the gap between informative clinical diagnostic data and simple class labels. Our method demonstrates state-of-the-art performance on multiple chest X-ray and mammography datasets compared with various baselines. The proposed parameter efficient framework can reduce the total trainable model size by 39\% and reduce the trainable language model to only 4\% compared with the current BERT encoder.
    \footnote{The official implementation is available at \url{https://github.com/XYPB/CLEFT}.}

\keywords{Deep Learning \and Multi-Modal \and Contrastive Learning \and Chest X-ray \and Mammography}
\end{abstract}

\begin{figure}[ht]
    \centering
    \includegraphics[width=1.0\textwidth]{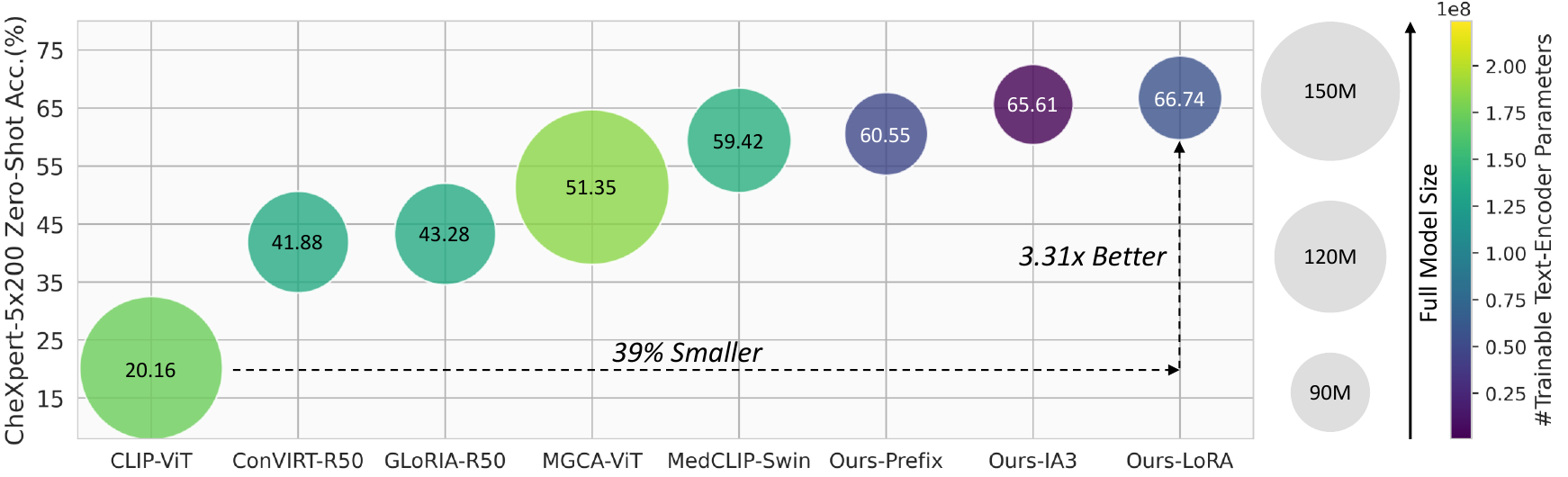}
    \caption{{\bf Zero-shot Performance on CheXpert-5x200.} We compare the performance of our method with multiple baselines on zero-shot CheXpert-5x200 classification. For each model, the diameter denotes the total number of trainable parameters, the color shows the number of trainable text encoder parameters, and the number reports accuracy. Our method outperforms all baselines with better parameter efficiency. }
    \label{fig:teaser}
\end{figure}
\mysection{Introduction}

Contrastive learning~\cite{he2019moco} has emerged as a pivotal paradigm in deep learning due to its ability to construct a robust feature space and generalize well to downstream tasks. Traditional contrastive learning methods~\cite{he2019moco,chen2020mocov2,chen2020simple,oquab2023dinov2} focus on building positive pairs from different views of the same input while distinguishing it from other data in the feature space. 
Such a contrastive paradigm allows the model to learn a robust representation when an exhaustive amount of training data is provided. 
Recently, Contrastive Language-Image Pre-training (CLIP)~\cite{radford2021learning}, which utilizes both visual and textual data insights, has extended contrastive learning to multi-modality data and reaped immense benefits from advancements in language models~\cite{radford2019language,brown2020language,sun2023eva}. CLIP supervises the image encoder with a language encoder trained simultaneously. By contrasting the features learned from both the image and language encoder, the CLIP model learns to bridge the textual and visual data in the high-dimensional embedding space, allowing it to use knowledge learned from the language model to guide visual encoder training. A properly pre-trained contrastive visual encoder can be adapted to multiple downstream tasks with a minimum amount of labeled data required. 

However, difficulties arise when adapting CLIP from the natural image-text pair to the medical domain, where access to data can be severely restricted by various factors, including security and privacy concerns, difficulty in obtaining expert annotations, and expensive imaging. The limited image-text pairs in the medical domain constrain the potential of CLIP models trained from scratch. While existing medical CLIP methods~\cite{huang2021gloria,zhang2022contrastive,wang2022multi,wang2022medclip} use a pre-trained BERT language model~\cite{alsentzer-etal-2019-publicly}, its limited model size constrained the expression ability in the embedding space and further constrained the pre-training capability. 
Also, the common approach of handcrafting textual prompts for CLIP~\cite{radford2021learning,huang2021gloria,wang2022medclip} leads to a lack of diversity in text training prompts, which can result in the catastrophic forgetting phenomenon~\cite{luo2023empirical} in the text encoder and limit the model's performance. 

In light of these challenges, we introduce a novel language-image Contrastive Learning method with Efficient LLM and prompt context Fine-Tuning (CLEFT) to boost overall performance. We leverage the strengths of vast pre-trained large language models (LLMs) and visual models, adapting them to the medical domain to counterbalance the scarcity of medical data and address the constraints due to language model size. Our work is the first in the realm of medical imaging to scale up the language model in the language-image pre-training to the billion parameter level, leading to greatly improved model generalization ability. Further, our approach is parameter efficient, solely focusing on optimizing smaller adaptation layers~\cite{hu2021lora,li2021prefixtuning} in the LLM, thus conserving GPU resources without compromising knowledge acquired from natural language. Our model reduces the total number of trainable parameters by 39\% with only 4\% of the trainable language model parameters compared to the vanilla CLIP~\cite{radford2019language} (\cref{fig:teaser}). In addition, our method learns a context-based prompt with prompt fine-tuning \cite{zhou2022learning} to mitigate the bias introduced by the undiversified hand-crafted prompt and generalize to different unseen data. The proposed method shows state-of-the-art performance on two public chest X-ray datasets \cite{irvin2019chexpert,rsna-pneumonia-detection-challenge} and a public mammography dataset~\cite{jeong2023emory}. 

\mysection{Methods}
\begin{figure}[t]
    \centering
    
    \includegraphics[width=1.0\textwidth]{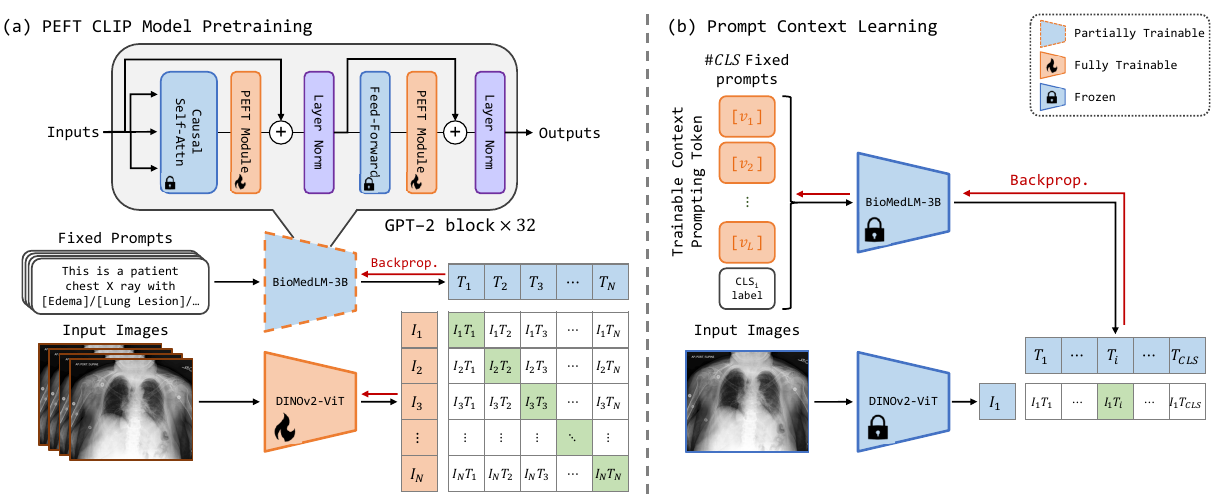}
    \caption{{\bf Proposed Method Framework.}  
    (a) Language-image contrastive learning with an LLM by utilizing PEFT. Fixed handcrafted prompts are used in this stage. (b) Prompt context learning with the pre-trained image and text encoder via classification.\vspace{-2mm}}
    \label{fig:method}
\end{figure}

The proposed CLEFT framework is in \cref{fig:method}. 
We first efficiently incorporate an LLM into the CLIP~\cite{radford2021learning} framework. We then train the learnable prompt context with frozen pre-trained text and visual encoders to further improve generalization. 

\mysubsection{Boosting CLIP with an LLM} 

\subsubsection{Contrastive Language-Image Pre-training.} The conventional CLIP~\cite{radford2021learning} framework includes a vision encoder and a text encoder with the corresponding projection head that encodes the image-text pair $(x_{I_i}, x_{T_i})$ sampled from the training data to feature $(I_i, T_i)$ (\cref{fig:method}(a)). The projection head maps the embedding from two different modalities into the same feature space and therefore allows the model to bridge two modalities. The multi-modal contrastive learning optimizes both image-to-text and text-to-image InfoNCE~\cite{he2019moco} loss symmetrically:
\begin{equation}
    \mathcal{L} = -\frac{1}{2N}\Big[\underbrace{\sum^{N}_{i=0}\log{\frac{\exp(I_i\cdot T_i / \tau)}{\sum^N_{j\neq i} \exp(I_i\cdot T_j / \tau)}}}_{\mathcal{L}_{I2T}} + \underbrace{\sum^{N}_{i=0}\log{\frac{\exp(T_i\cdot I_i / \tau)}{\sum^N_{j\neq i} \exp(T_i\cdot I_j / \tau)}}}_{\mathcal{L}_{T2I}}\Big],
    \label{eq:loss}
\end{equation}
where $N$ is the number of samples within a batch, and $\tau$ is the learnable softmax temperature. Optimizing the loss will reduce the cosine distance between paired image and text features while repelling unpaired samples. 
However, proper contrastive learning requires a large amount of training data as well as a large negative sample size (\ie, batch size), and it is hard to gather large datasets in the medical domain and GPU resources are limited. Thus, we propose to replace the vanilla encoders trained from scratch with fully pre-trained models to alleviate this issue. Initializing the text encoder with a pre-trained model provides higher-quality supervision since the pre-trained model already provides a robust feature space~\cite{chen2023contrastive}, which simplifies the contrastive learning procedure. 

\subsubsection{PEFT LLM as Text Encoder.} To further explore the potential of the CLIP framework, we take advantage of a medical LLM that was pre-trained with a large amount of text data. We use a GPT-2-based causal language model as our text encoder rather than a BERT-based language model, which was widely used in previous medical CLIP models~\cite{zhang2022contrastive,huang2021gloria,wang2022multi,wang2022medclip}, since the causal LLM has shown a better capability as it scales up to over a billion parameters~\cite{brown2020language}. A stronger text encoder allows the model to embed the input into a more robust feature space with less training. However, LLMs are more likely to overfit on undiversified text data given their strong expression ability. To avoid this issue and maintain the robust pre-trained knowledge within the LLM efficiently, we introduce the parameter-efficient fine-tuning (PEFT) module to the frozen LLM (\cref{fig:method}(a)), where a small set of trainable parameters are injected into each transformer block, adjusting the original output of the attention layers slightly. This reduces the number of trainable parameters during training to no more than 1\% of the full large language model size. Common PEFT methods like LoRA~\cite{hu2021lora} and IA3~\cite{liu2022fewshot} either adjust the attention output with low-rank bottleneck matrices or scale the key, query, and value outputs. Also, prefix fine-tuning~\cite{li2021prefixtuning} introduces extra trainable prefix tokens to the language model to influence the behavior of the attention layers. The nature of these PEFT methods ensures the fine-tuned output will not deviate too much from the original model and helps avoid the catastrophic forgetting phenomenon~\cite{luo2023empirical}. To further merge the domain gap between pre-training text data and CLIP prompt, we unlock the LLM's embedding layer and update the corresponding token embedding during pre-training.

\subsubsection{CLIP as Knowledge Distillation.} 
We further argue that contrastive language-image learning with a pre-trained text encoder can be viewed as a knowledge distillation process, where the numerator in \cref{eq:loss} minimizes the distance between $I_i$ and $T_i$. Given that the language encoder is largely frozen, the language encoder serves more as a teacher model since it already has a well-established embedding space that can distinguish between different samples, and its output only changes slightly during training. The fully optimized visual encoder then focuses on aligning its output to the language model's embedding space, which serves as the student model. This allows us to distill the knowledge within the well-pre-trained language model into a generally much smaller vision model and, therefore, further improve its performance. Additionally, since the ``teacher model'' is still optimized during pre-training, the negative pairs are still necessary.

\subsubsection{Model Architecture.} We choose GPT-2~\cite{radford2019language} with 32 causal transformer blocks as the text encoder and the ViT-Base~\cite{dosovitskiy2020image} with a patch size of 14 as the visual encoder (\cref{fig:method}(a)). Similar to the original CLIP~\cite{radford2021learning}, we use a randomly initialized linear projection layer to map the embeddings from each encoder to a unified embedding space with the same size. For the text encoder, we use the output embedding of the first \texttt{[EOS]} token since the model is causal and this token encodes all the information from the input. For the PEFT module, we experiment with LoRA~\cite{hu2021lora}, IA3~\cite{liu2022fewshot}, and prefix fine-tuning~\cite{li2021prefixtuning}. For the visual encoder, we use the averaged embedding of all visual tokens from ViT~\cite{dosovitskiy2020image} as the visual embedding. We remove the last layer norm for better training stability.

\mysubsection{Learning the Context-Based Prompt} 
To further address the issue of the lack of diversity in the hand-crafted prompts, we introduce a second stage of training that only optimizes a learnable context-based token of length $L$ (\cref{fig:method}(b)). After pre-training, we freeze both encoders and replace the original hand-crafted prompt with a series of trainable tokens that then feed into the language model. The same context-based prompt tokens are used from all classes, which ensures the generalization ability of these tokens. Different from the pre-training stage, we optimize the trainable context tokens with a zero-shot classification cross-entropy loss. This allows the prompt tokens to adapt to different classes evenly and avoid the potential shortcut issue. We further initialize these tokens with the embedding of the original hand-crafted caption. If $L$ is longer than the original caption, we instead initialize the first few tokens according to the random uniform distribution.

\mysection{Experiments}
\label{sec:experiment}

\mysubsection{Datasets}

We evaluate our CLEFT model on two major applications in medical imaging, chest X-ray and mammography.
We use the \textbf{CheXpert-1.0}~\cite{irvin2019chexpert} for pre-training following GLoRIA~\cite{huang2021gloria}. The dataset has 223,415 images from 65,240 patients with corresponding class labels for 14 different classes. We only use frontal chest radiographs for consistency. We leave out 5\% of data for validation. For evaluation, we use the in-domain \textbf{CheXpert-5x200}~\cite{huang2021gloria} and out-of-domain \textbf{RSNA}~\cite{rsna-pneumonia-detection-challenge} datasets. The CheXpert-5x200~\cite{huang2021gloria} is a multi-class classification subset of CheXpert-1.0 with 5 different classes and 200 images in each class. These images are removed from the training set. The RSNA~\cite{rsna-pneumonia-detection-challenge} provides a collection of pneumonia and non-pneumonia images for binary classification. Following MedCLIP~\cite{wang2022medclip}, we sample a 1:1 subset with 8,486 training and 3,538 testing images for this dataset. For the mammography data, we use the \textbf{EMBED}~\cite{jeong2023emory} dataset which contains 364,515 2D mammograms with both BI-RADS \cite{Sickles2013} and breast density labels. We split the data into 70\%/10\% for training/validation. We evaluate our model on both BI-RADS and density prediction tasks with two balanced subsets sampled from the 20\% remaining data.
The BI-RADS classification test set contains 7 classes with 200 samples per class. The density classification test set contains 4 classes each with 500 samples.

\mysubsection{Baselines and Evaluation Metrics}
We compare multiple state-of-the-art baselines. To demonstrate the effectiveness of the CLIP pre-training, we compare with the same \textbf{ViT}~\cite{oquab2023dinov2} model with random initialization and Image-Net~\cite{deng2009imagenet} pre-training. We further compare our model with conventional \textbf{CLIP}~\cite{radford2021learning}, ResNet50~\cite{he2016deep} based medical CLIP method \textbf{ConVIRT}~\cite{zhang2022contrastive} and \textbf{GLoRIA}~\cite{huang2021gloria}. We also compare with recent medical CLIP baselines including \textbf{MGCA}~\cite{wang2022multi}, \textbf{MRM}~\cite{zhou2023advancing}, and \textbf{MedCLIP}~\cite{wang2022medclip} with Swin-Transformer~\cite{liu2021swin}. We choose these baselines as they provide either their pre-trained model or full training code. However, the ConVIRT~\cite{zhang2022contrastive} model does not provide a pre-trained model, so we report results directly drawn from Wang~\etal\cite{wang2022medclip}.

We evaluate all models under zero-shot, linear-probing, and full fine-tuning settings. 
We report accuracy for zero-shot classification and both accuracy and area under the receiver operating characteristic curve (AUC) for the two fine-tuning settings. We further evaluate the data efficiency during full fine-tuning of our model and compare the model size.

\mysubsection{Implementation Details} 

We choose BioMedLM-3B~\cite{biomedLM2023} and DiNOv2~\cite{oquab2023dinov2} to initialize our encoders.
During the pre-training, we use a batch size of 72 and learning rate of $4\times10^{-5}$ for 40,000 steps. We use the cosine annealing scheduler with a 4,000-step linear warm-up and AdamW~\cite{loshchilov2017decoupled} optimizer with weight decay of $0.2$. We select the model with the smallest validation loss as the final model. During prompt context learning, we use a batch size of 36 and a learning rate of $1\times10^{-3}$ for 4,000 steps. Prompt context length is set to $L=30$. We use the same scheduler with a 1,000-step warm-up and SGD optimizer. For the PEFT strategy, we experiment with LoRA~\cite{hu2021lora}, IA3~\cite{liu2022fewshot} and Prefix-tuning~\cite{li2021prefixtuning}. For linear probing and full fine-tuning, we optimize cross-entropy loss using a batch size of 36 and learning rate of $5\times10^{-4}$ for 8,000 steps with weight decay of $1\times 10^{-3}$. All models are trained with BFloat-16-mix precision with 2 NVIDIA A5000 GPUs using PyTorch.

\begin{table}[t]
    \caption{{\bf Main Evaluation Results.} We evaluate the proposed method with multiple baselines on the CheXpert-5$\times$200~\cite{irvin2019chexpert} and RSNA~\cite{rsna-pneumonia-detection-challenge} datasets. We evaluate zero-shot (ZS) classification, linear-probing (LP), and full-finetuning (FT) tasks. All values are percentages. We highlight the top result in bold and the second-best with an underline. }\label{tab:main}
    \centering
    \resizebox{\textwidth}{!}
    {
    \begin{tabular}{@{}lccccc|ccccc@{}}
    \toprule
    \multicolumn{1}{c}{\multirow{2}{*}{\textbf{Method}}} & \multicolumn{5}{c|}{\textbf{CheXpert} $\bm{5\times200}$~\cite{irvin2019chexpert}} & \multicolumn{5}{c}{\textbf{RSNA}~\cite{rsna-pneumonia-detection-challenge}} \\ \cmidrule(l){2-11} 
    \multicolumn{1}{c}{} & ~ZS-Acc~ & ~LP-Acc~ & ~LP-AUC~ & ~FT-Acc~ & ~FT-AUC~ & ~ZS-Acc~ & ~LP-Acc~ & ~LP-AUC~ & ~FT-Acc~ & ~FT-AUC~ \\ \midrule
    Random-ViT~\cite{oquab2023dinov2} & - & 20.62 & 57.85 & 20.32 & 63.68 & - & 65.15 & 72.22 & 72.70 & 79.89 \\
    ImageNet-ViT~\cite{oquab2023dinov2} & - & 38.54 & 75.44 & 56.46 & 85.71 & - & 74.96 & 83.72 & 77.44 & 85.24 \\ \midrule
    CLIP-ViT~\cite{radford2021learning} & 20.16 & 35.34 & 65.14 & 44.84 & 77.59 & 49.89 & 69.47 & 75.90 & 77.08 & 83.53 \\
    ConVIRT-R50~\cite{zhang2022contrastive} & ~41.88$^*$ & ~47.70$^*$ & - & - & - & ~47.31$^*$ & ~78.46$^*$ & - & - & - \\
    GLoRIA-R50~\cite{huang2021gloria} & ~43.28$^*$ & 51.65 & 83.15 & 57.56 & 87.11 & ~33.06$^*$ & 76.77 & 86.53 & 78.55 & 87.15 \\
    MGCA-ViT$^\dagger$~\cite{wang2022multi} & 51.35 & 26.63 & 63.94 & 56.96 & 86.33 & {\ul 69.90} & 75.35 & 85.63 & 79.79 & 88.11 \\
    MRM-ViT~\cite{zhou2023advancing} & - & 58.26 & 84.48 & 56.56 & 87.41 & - & 76.43 & 85.48 & 78.77 & 86.63 \\ 
    MedCLIP-Swin$^\dagger$~\cite{wang2022medclip}  & ~59.42$^*$ & 54.55 & 85.75 & 57.46 & {\ul 87.85} & ~\textbf{74.47}$^*$ & {\ul 79.26} & 88.26 & 78.80 & 87.36 \\ \midrule
    Ours-Prefix~ & 60.55 & {\ul 61.56} & {\ul 87.20} & {\ul 61.16} & 87.73 & 64.07 & 78.75 & {\ul 88.30} & 79.34 & 88.52 \\
    Ours-IA3 & {\ul 65.61} & 60.16 & 86.48 & 61.06 & 86.81 & 64.08 & 78.80 & 87.74 & {\ul 79.99} & {\ul 88.59} \\
    Ours-LoRA & \textbf{66.74} & \textbf{63.46} & \textbf{87.76} & \textbf{63.96} & \textbf{88.22} & 64.93 & \textbf{79.40} & \textbf{88.34} & \textbf{80.36} & \textbf{88.72} \\ \bottomrule
    \multicolumn{11}{l}{$^*$ Result directly drawn from Wang~\etal\cite{wang2022medclip}} \\
    \multicolumn{11}{l}{$^\dagger$ Method pre-trained with a different dataset with 2$\times$ greater size}
    \end{tabular}
    }
    

\end{table}

\mysubsection{Main Results}
\subsubsection{Zero-shot Classification.} As shown in \cref{tab:main}, our model with LoRA~\cite{hu2021lora} outperforms other baselines with 7\% improvement on the CheXpert-5x200~\cite{irvin2019chexpert} dataset and with the smallest number of trainable language parameters and overall small trainable model size (\cref{fig:teaser}, \cref{tab:model_size}). Note that while our model falls behind MGCA~\cite{wang2022multi} and MedCLIP~\cite{wang2022medclip} on the RNSA~\cite{rsna-pneumonia-detection-challenge} evaluation, these two baselines were pre-trained with a different, larger dataset with twice the size of CheXpert-1.0~\cite{irvin2019chexpert}. Compared with the other baselines that were pre-trained with the same data, our method performs best on the out-of-domain RSNA data.

\subsubsection{Linear probing.} Under the linear probing condition, our model with LoRA~\cite{hu2021lora} achieves the best performance on both datasets (\cref{tab:main}). 
We highlight the 5\% gap in the CheXpert-5x200~\cite{irvin2019chexpert} experiment. 
This indicates our model has a more robust embedding space that can distinguish input data even without task-specific fine-tuning. We further note that with linear probing, our method now surpasses MGCA~\cite{wang2022multi} and MedCLIP~\cite{wang2022medclip} on RSNA, even with less pre-training data. 

\subsubsection{Full Fine-tuning.}
Our model outperforms all other baselines when the model is fully fine-tuned (\cref{tab:main}). Our model shows impressive improvement on the CheXpert~\cite{irvin2019chexpert} dataset and also beats other baselines on the RSNA~\cite{rsna-pneumonia-detection-challenge} dataset. We suggest it is the proposed PEFT language model that provides a better quality of supervision. Meanwhile, the vastly pre-trained encoders allow the model to properly adapt to out-of-domain tasks. We further evaluate each model's data efficiency with different ratios of training data in \cref{tab:chest_ratio} and \cref{tab:rsna_ratio}. A more robust pre-trained model should be able to generalize easily to the target task even with a small amount of training data. Notably, our model also outperforms the baselines even with much less training data. We highlight that for RSNA, our accuracy drops by $<3\%$ when using $1\%$ compared to $100\%$ of the training data. 

\subsubsection{Mammography Evaluation.} We report the performance of our model with LoRA~\cite{hu2021lora} and baselines on the EMBED mammography data in \cref{tab:breast}. Our model clearly surpasses the compared baselines with a considerable gap. These initial results on the mammography dataset suggest the proposed model has the potential to be applied to other medical domains.

\begin{table}[t]

    \begin{minipage}[t]{0.49\textwidth}
        \centering
        \caption{{\bf EMBED Evaluation.} All values are percentages.}
        \label{tab:breast}
        \resizebox{\textwidth}{!}
        {
        \begin{tabular}{@{}lccc|ccc@{}}
        \toprule
        \multicolumn{1}{c}{\multirow{2}{*}{\textbf{Method}}} & \multicolumn{3}{c|}{\textbf{BI-RADS}} & \multicolumn{3}{c}{\textbf{Density}} \\ \cmidrule(l){2-7} 
        \multicolumn{1}{c}{} & ~ZS-Acc~ & ~LP-Acc~ & ~FT-Acc~ & ~ZS-Acc~ & ~LP-Acc~ & ~FT-Acc~ \\ \midrule
        Random~\cite{oquab2023dinov2} & - & 17.86 & 17.43 & - & 34.65 & 54.45 \\
        ImageNet~\cite{oquab2023dinov2} & - & 18.29 & 20.21 & - & 56.95 & 67.45 \\ \midrule
        CLIP~\cite{radford2021learning} & 20.93 & 16.57 & 21.21 & 70.45 & 36.70 & 50.50 \\
        Ours-LoRA & \textbf{33.29} & \textbf{30.86} & \textbf{23.79} & \textbf{74.40} & \textbf{74.80} & \textbf{74.95} \\ \bottomrule
        \end{tabular}
        }
        
    \end{minipage}%
    \hfill
    \begin{minipage}[t]{0.49\textwidth}
        \centering
        \caption{{\bf Ablation Evaluation.} All values are percentages.}
        \label{tab:abalation}
        \resizebox{.98\textwidth}{!}
        {
        \begin{tabular}{@{}lcc|c@{}}
        \toprule
        \multicolumn{1}{c}{\multirow{2}{*}{\textbf{Method}}} & \multirow{2}{*}{\textbf{LM \#Param.}} & \multicolumn{2}{c}{\textbf{Zero-shot Acc.}} \\ \cmidrule(l){3-4} 
         &  & CheXpert~$5\times200$~\cite{irvin2019chexpert} & RSNA~\cite{rsna-pneumonia-detection-challenge} \\ \midrule
        Ours-LoRA & 2.6M & {\ul 66.74}\diff{\pm 0.00} & {\ul 64.63}\diff{\pm 0.00} \\ \midrule
        - Prompt FT. & 2.6M & 63.46\diff{-3.28} & 57.15\diff{-7.48} \\
        - Freeze LM & 0.0M & 64.44\diff{-2.23} & 65.51\diff{+0.88} \\
        + Full FT. LM & 2.7B & \textbf{67.03\diff{+0.29}} & \textbf{67.58}\diff{+2.95} \\\bottomrule
        \end{tabular}\label{tab:ablation}
        }
        
    \end{minipage}

\end{table}

\begin{figure}[t]
    \centering
    \includegraphics[width=.95\textwidth]{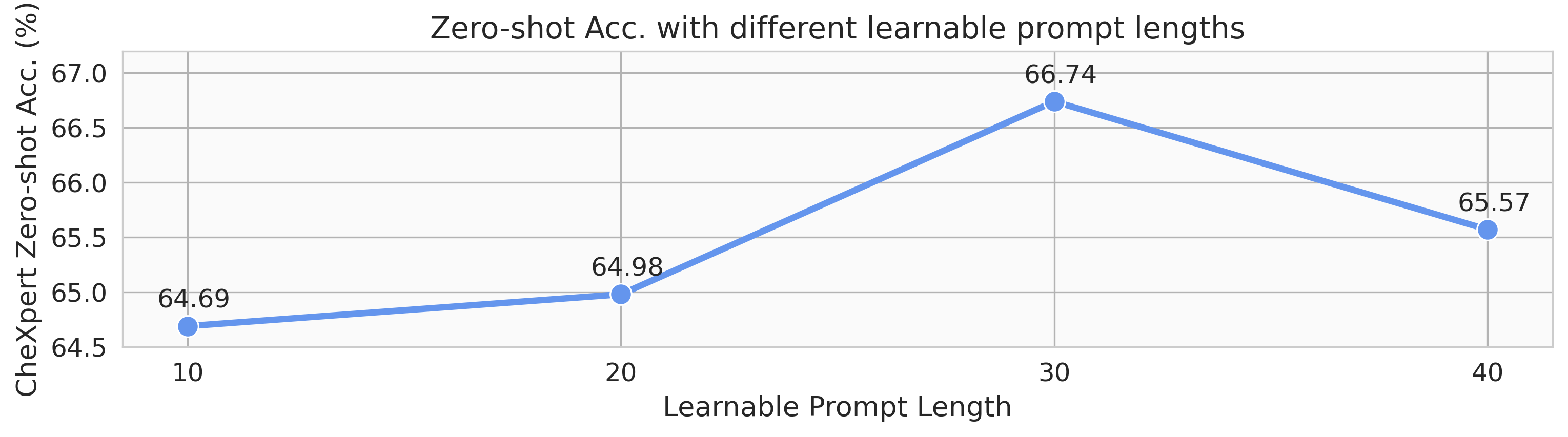}
    \vspace{-2mm}
    \caption{\textbf{Zero-shot Accuracy vs. Number of Trainable Prompt Tokens.}}
    \label{fig:prompt}
\end{figure}

\mysubsection{Ablation Experiments}
\label{sec:ablation}
Results of ablation experiments with our method are presented in \cref{tab:ablation}. Without the second stage prompt fine-tuning, the accuracy drops by $\sim$3\% for both datasets. Using a fully frozen language model harms performance on in-domain data but improves performance with out-of-domain data. The fully fine-tuned model improves the performance even with a much smaller batch size; however, this improvement comes with the cost of $\sim$4 times more GPU memory cost and only $1/20$ batch size with 2 times longer training. We also evaluate the influence of using different lengths for the prompt context (\cref{fig:prompt}), and we choose $L=30$ as our best model in the other evaluations. Increasing the prompt length does not always improve the performance according to this experiment.

\mysection{Discussion and Conclusion}

We propose a novel contrastive language-image pre-training framework for medical images, called CLEFT, which incorporates a pre-trained medical LLM, parameter-efficient fine-tuning for the LLM, and prompt context learning. The proposed method successfully outperforms multiple baselines in the experiments conducted on three public datasets. Furthermore, the proposed method greatly reduces the training cost of incorporating an LLM while maintaining robust performance. We provide a new view that treats the existing CLIP task as knowledge distillation from the LLM. The proposed framework can be easily merged with other related CLIP methods like MedCLIP~\cite{wang2022medclip} to further improve the performance.

While the proposed method demonstrated enhanced performance, additional methodological improvements should be explored, \eg, different initialization methods, more robust prompt finetuning methods that can adapt to the medical image domain, and generating more diverse and detailed prompting during training. Finally, more experiments including text-image retrieval and benchmarking of different visual encoders and LLM models should be conducted to further evaluate the methods. 

\subsubsection{Acknowledgments} This work was supported by NIH grant R21EB032950.

\subsubsection{Disclosure of Interests} The authors have no competing interests in this work and other related research.

\bibliographystyle{splncs04}
\bibliography{clip}
\setcounter{tocdepth}{1}

\clearpage

\setcounter{page}{1}
\setcounter{section}{0}
\setcounter{figure}{0}
\setcounter{table}{0}
\renewcommand{\thepage}{S\arabic{page}} 
\renewcommand{\thesection}{S\arabic{section}}
\renewcommand{\thefigure}{S\arabic{figure}}%
\renewcommand{\thetable}{S\arabic{table}}%

\section{Supplementary Material}
\label{sec:supplement}

\begin{table}[h]
\centering
\caption{{\bf Data Efficiency Evaluation on CheXpert~\cite{irvin2019chexpert}.} We evaluate the full-finetuning performance of the proposed method with multiple baselines on the CheXpert-5$\times$200~\cite{irvin2019chexpert} datasets with different ratios of training data (1\%/10\%/100\%). A more robust pre-trained model should be able to generalize easily to the target task even with a small amount of training data. We highlight the top result in bold and the second-best with an underline.}

\resizebox{\textwidth}{!}
{
\begin{tabular}{@{}l|cc|cc|cc@{}}
\toprule
\multicolumn{1}{c}{\multirow{3}{*}{\textbf{Method}}} & \multicolumn{6}{c}{\textbf{CheXpert} \bm{$5\times200$}~\cite{irvin2019chexpert}} \\ \cmidrule(l){2-7} 
\multicolumn{1}{c}{} & \multicolumn{2}{c|}{1\%} & \multicolumn{2}{c|}{10\%} & \multicolumn{2}{c}{100\%} \\ \cmidrule(l){2-7} 
\multicolumn{1}{c}{} & ~FT-Acc~ & ~FT-AUC~ & ~FT-Acc1~ & ~FT-AUC~ & ~FT-Acc~ & ~FT-AUC~ \\ \midrule
Random-ViT~\cite{oquab2023dinov2} & 21.42 & 58.81 & 20.02 & 54.96 & 20.32 & 63.68 \\
ImageNet-ViT~\cite{oquab2023dinov2} & 35.54 & 67.68 & 45.15 & 75.52 & 56.46 & 85.71 \\ \midrule
CLIP-ViT-BERT~\cite{radford2021learning} & 23.82 & 62.78 & 40.74 & 71.20 & 44.84 & 77.59 \\
GLoRIA-R50~\cite{huang2021gloria} & 48.45 & 77.08 & 53.05 & 83.34 & 57.56 & 87.11 \\
MGCA-ViT~\cite{wang2022multi} & 45.55 & 75.85 & 54.65 & 82.54 & 56.96 & 86.33 \\
MRM-ViT~\cite{zhou2023advancing} & 49.35 & 78.84 & {\ul 55.86} & {\ul 86.06} & 56.56 & 87.41 \\
MedCLIP-Swin~\cite{wang2022medclip}  & {\ul 50.65} & \textbf{80.60} & 52.95 & 82.91 & 57.46 & {\ul 87.85} \\ \midrule
Ours-Prefix~\cite{li2021prefixtuning} & 45.35 & 77.17 & 55.46 & 83.67 & {\ul 61.16} & 87.73 \\
Ours-IA3~\cite{liu2022fewshot} & 46.65 & 75.69 & 53.65 & 82.23 & 61.06 & 86.81 \\
Ours-LoRA~\cite{hu2021lora} & \textbf{51.15} & {\ul 80.10} & \textbf{56.96} & \textbf{86.28} & \textbf{63.96} & \textbf{88.22} \\ \bottomrule
\end{tabular}
}
\label{tab:chest_ratio}
\end{table}
\begin{table}[ht]
\centering
\caption{{\bf Data Efficiency Evaluation on RSNA~\cite{rsna-pneumonia-detection-challenge}.} We evaluate the full-finetuning performance of the proposed method with multiple baselines on the out-of-domain RSNA~\cite{rsna-pneumonia-detection-challenge} datasets with different ratios of training data (1\%/10\%/100\%). A more robust pre-trained model should be able to generalize easily to the target task even with a small amount of training data. Our accuracy drops by <3\% when using 1\% training data compared to 100\% training data. We highlight the top result in bold and the second-best with an underline.}

\resizebox{\textwidth}{!}
{
\begin{tabular}{@{}l|cc|cc|cc@{}}
\toprule
\multicolumn{1}{c}{\multirow{3}{*}{\textbf{Method}}} & \multicolumn{6}{c}{\textbf{RSNA}~\cite{rsna-pneumonia-detection-challenge}} \\ \cmidrule(l){2-7} 
\multicolumn{1}{c}{} & \multicolumn{2}{c|}{1\%} & \multicolumn{2}{c|}{10\%} & \multicolumn{2}{c}{100\%} \\ \cmidrule(l){2-7} 
\multicolumn{1}{c}{} & ~FT-Acc~ & ~FT-AUC~ & ~FT-Acc~ & ~FT-AUC~ & ~FT-Acc~ & ~FT-AUC~ \\ \midrule
Random-ViT~\cite{oquab2023dinov2} & 62.15 & 66.23 & 71.76 & 78.65 & 72.70 & 79.89 \\
ImageNet-ViT~\cite{oquab2023dinov2} & 71.71 & 78.11 & 76.09 & 83.97 & 77.44 & 85.24 \\ \midrule
CLIP-ViT-BERT~\cite{radford2021learning} & 66.48 & 71.45 & 76.09 & 76.64 & 77.08 & 83.53 \\
GLoRIA-R50~\cite{huang2021gloria} & 74.79 & 82.13 & 76.29 & 83.28 & 78.55 & 87.15 \\
MGCA-ViT~\cite{wang2022multi} & 74.22 & 82.13 & 76.37 & 83.02 & 79.79 & 88.11 \\
MRM-ViT~\cite{zhou2023advancing} & 72.98 & 80.93 & 76.37 & 84.70 & 78.77 & 86.63 \\ 
MedCLIP-Swin~\cite{wang2022medclip}  & 75.55 & 83.41 & 77.33 & {\ul 85.79} & 78.80 & 87.36 \\ \midrule
Ours-Prefix~\cite{li2021prefixtuning} & {\ul 76.40} & {\ul83.99} & {\ul 78.38} & 85.55 & 79.34 & 88.52 \\
Ours-IA3~\cite{liu2022fewshot} &74.52 & 82.26 & 77.59 & 85.47 & {\ul 79.99} & {\ul 88.59} \\
Ours-LoRA~\cite{hu2021lora} & \textbf{77.53} & \textbf{84.81} & \textbf{78.38} & \textbf{86.22} & \textbf{80.36} & \textbf{88.72} \\ \bottomrule
\end{tabular}
}
\label{tab:rsna_ratio}
\end{table}

\begin{table}[ht]
\centering
\caption{{\bf Model Trainable Parameter Size.} We list the total trainable model size, trainable vision encoder size, and trainable language model size for each baseline and our method.}

\resizebox{\textwidth}{!}
{
\begin{tabular}{@{}lccc@{}}
\toprule
\textbf{Model Name} & ~\textbf{Total Trainable Size}~ & ~\textbf{Vision Encoder Size}~ & ~\textbf{Language Model Size}~ \\ \midrule
ViT-B-14~\cite{oquab2023dinov2} & 90.42M & 90.42M & - \\ \midrule
CLIP-ViT-BERT~\cite{radford2021learning} & 153.59M & 90.42M & 63.16M \\
ConVIRT-R50-BERT~\cite{zhang2022contrastive} & 108.13M & 25.08M & 83.05M \\
GLoRIA-R50-BERT~\cite{huang2021gloria} & 108.14M & 25.08M & 83.05M \\
MGCA-ViT~\cite{wang2022multi} & 168.86M & 85.80M & 83.05M \\
MRM-ViT~\cite{zhou2023advancing} & 168.85M & 85.79M & 83.05M \\
MedCLIP-Swin-BERT~\cite{wang2022medclip} & 110.98M & 27.91M & 83.05M \\ \midrule
Ours-ViT-GPT2(Prefix)~\cite{li2021prefixtuning} & 93.04M & 90.42M & 2.62M \\
Ours-ViT-GPT2(IA3)~\cite{liu2022fewshot} & 90.99M & 90.42M & 0.57M \\
Ours-ViT-GPT2(LoRA)~\cite{hu2021lora} & 93.70M & 90.42M & 3.27M \\ \bottomrule
\end{tabular}
}
\label{tab:model_size}
\end{table}

\end{document}